\documentclass[runningheads]{llncs}
\usepackage[T1]{fontenc}
\usepackage{graphicx}
\usepackage{tabularx}
\usepackage{subcaption}
\usepackage{cite}
\begin{document}
\title{Bayesian Networks for Named Entity Prediction in Programming Community Question Answering}
%
%
\author{Alexey Gorbatovski\orcidID{0000-0003-3705-0047} \and \\
Sergey Kovalchuk\orcidID{0000-0001-8828-4615}}
\authorrunning{Gorbatovski and Kovalchuk}
%
\institute{ITMO University, Saint-Petersburg, Russia \\
\email{gorbatovski@itmo.ru}, \email{kovalchuk@itmo.ru}}
\maketitle              
\begin{abstract}

Within this study, we propose a new approach for natural language processing using Bayesian networks to predict and analyze the context and how this approach can be applied to the Community Question Answering domain. We discuss how Bayesian networks can detect semantic relationships and dependencies between entities, and this is connected to different score-based approaches of structure-learning. We compared the Bayesian networks with different score metrics, such as the BIC, BDeu, K2 and Chow-Liu trees. Our proposed approach out-performs the baseline model at the precision metric. We also discuss the influence of penalty terms on the structure of Bayesian networks and how they can be used to analyze the relationships between entities. In addition, we examine the visualization of directed acyclic graphs to analyze semantic relationships. The article further identifies issues with detecting certain semantic classes that are separated in the structure of directed acyclic graphs. Finally, we evaluate potential improvements for the Bayesian network approach.

\end{abstract}
\section{Introduction}

Increasing interest in natural language processing (NLP) presented automated solutions to different human problems, such as text classification, text summarization and generation either with quality comparable to human solutions \cite{khurana2022natural}. Still, there remains the problem of context addition. To solve this problem we aim to predict the entire context or basic entities to get the coherent and cohesive text meaning \cite{santhanam2020context}. Usually a part of the full text, such as the key words or some description, are available. On the one hand it is a more complicated task than a language modeling problem, because there are limits to extrapolating the context from a small part or a description \cite{pillutla2021mauve}. Furthermore, the semantic gap between the original text and what is recovered is not as unambiguous as in the summarization problem \cite{nallapati2016abstractive}.

This may also be applied to search insights by titles or recovery of text contents when the author is not available. Another application is feature extraction for better text generation or context reconstruction for dialogues \cite{yang2019end}. On the other hand, there are topics in the analysis and use of human code generation quality assessment \cite{kovalchuk-etal-2022-human} and community question answering (CQA), in which the Bayesian approach could prove a great tool. The CQA domain needs to emphasize information from questions or shorter titles, to generate answers more accurately. Such tasks mostly obtain good results by complex and sophisticated neural network architectures, such as LSTM \cite{santhanam2020context} or transformers \cite{10.1007/978-3-030-64580-9_32}. However, there are issues with this application of neural networks \cite{liu-etal-2020-understanding}. For example, well known GPT-like models used for text generation need huge amounts of textual data and time, and they are too complex for fine tuning \cite{Brown2020LanguageMA}.

In this paper, we present a Bayesian approach for context prediction. Bayesian networks (BNs) allow us to recover the meaning of a full text by knowing the conditional probability distributions (CPDs) of named entities. A named entity in our case is the class of one of the semantically meaningful words in the programming domain obtained as a result of named entity recognition (NER). These entities present informative units that carry information about the context.

Additionally, the directed acyclic graph (DAG) provided by BN show links between entity classes. In most cases, entities from the title part directly affect the appearance of the entity in question. Besides, it detects links between significant elements of the programming domain, such as code blocks with error names or class and function entity classes. In practice, because of probable errors of the NER model used to annotate the text content, there may be incorrect relationships, but in an ideal case, BN specify more precise relationships and give information about semantics and causal relationships.

\section{Methodology}

\begin{figure}
\includegraphics[width=\textwidth]{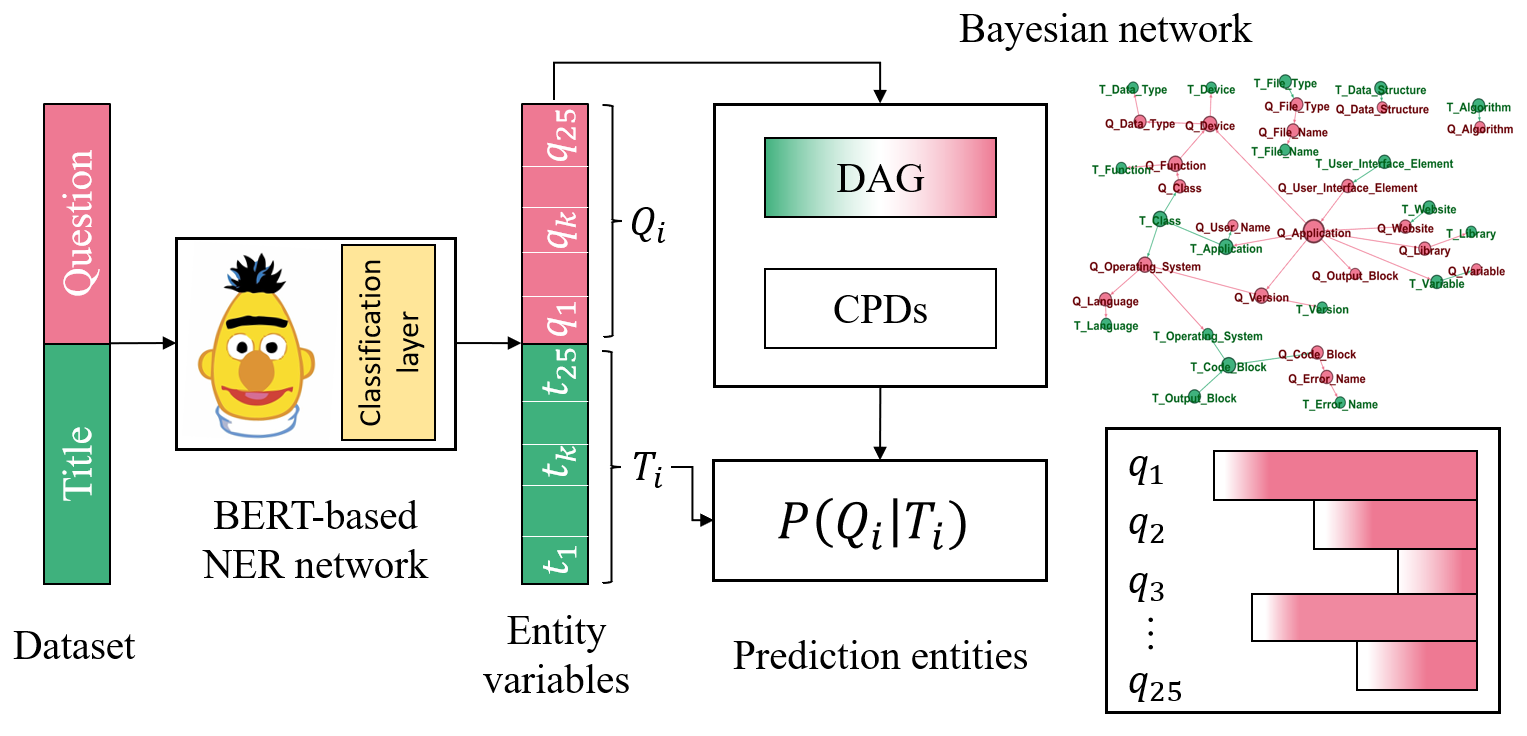}
\caption{The overall process of proposed BN approach} \label{fig1}
\end{figure}

In this section, we describe different components of the proposed BN approach for context prediction. Figure~\ref{fig1}, shows the overall process consists of several parts:
1) Semantic entity recognition by the NER model;
2) Learning the Bayesian network as a causal model;
3) Predicting and evaluating entities in question by title.

\subsection{Problem Statement}

As shown in Figure~\ref{fig1}, we need to predict the semantic meaningful classes of questions with BN as a multilabel classification problem. For this problem we have textual data, presented as vectors.

More formally, assume we are given two sets of Questions $Q = \langle Q_1, Q_2, \dots, Q_N \rangle$ and Titles $T = \langle T_1, T_2, ..., T_N \rangle$, where $N$ - is the number of samples in our dataset. For each title $T_i\in{T}$ we have $k=25$ dimension vector, $T_i= \langle t_1^i, t_2^i, ..., t_k^i \rangle$, where $t_k^i$ represents the $k_{th}$ entity class of the $i_{th}$ title and $t_k^i\in{\{0, 1\}}$, where $t_k^i = 0$ corresponds to the absence of the  $k_{th}$ class entity in title, and $t_k^i = 1$ corresponds to the existence of the $k_{th}$ class entity in title. For the questions it is the same, for each question $Q_i\in{Q}$ there are 25 dimension vectors, $Q_i= \langle q_1^i, q_2^i, ..., q_k^i \rangle$, where $q_k^i$ represents the $k_{th}$ entity class of the $i_{th}$ question and $q_k^i\in\{{0, 1\}}$, where $q_k^i = 0$ corresponds to the absence of the $k_{th}$ class entity in question, and $q_k^i = 1$ corresponds to existence of the $k_{th}$ class entity in question. We solve the multi label classification problem by predicting for each $i_{th}$ question its entity classes by $i_{th}$ titles entity classes.

\subsection{Dataset}

The dataset we use is based on 10\% of the Stack Overflow\footnote{\url{https://stackoverflow.com}} Q\&A 3 years ago\footnote{\url{https://www.kaggle.com/datasets/stackoverflow/stacksample}}. For the set of questions we apply the following filtering operations: select questions with tag "android", select questions with a length less than 200 words and related to the API Usage category proposed by Stefanie et al. \cite{beyer2020kind}. Moreover, we selected questions without links and images, because information from those types of content is unavailable for Bayesian networks. Thus, we received $N = 707$ pairs of title and question $(T_i, Q_i)$.

\subsection{Semantic Entities Recognition}

For extracting domain specific entities from text content we used the open source CodeBERT \cite{feng2020codebert} realization trained for the NER problem \cite{lample2016neural} on Stack Overflow data, since this is the most popular resource for programmers to find answers to questions. The model was tuned to detect 25 entity classes defined by Jeniya et al. \cite{tabassum-etal-2020-code}. They represent the following classes: ALGORITHM, APPLICATION, CLASS, CODE BLOCK, DATA STRUCTURE, DATA TYPE, DEVICE, ERROR NAME, FILE NAME, FILE TYPE, FUNCTION, HTML XML TAG, KEYBOARD IP, LANGUAGE, LIBRARY, LICENSE, OPERATING SYSTEM, ORGANIZATION, OUTPUT BLOCK, USER INTERFACE ELEMENT, USER NAME, VALUE, VARIABLE, VERSION, WEBSITE. Each class is domain specific and defines context semantics \cite{dash2008context}.

\begin{figure}
\includegraphics[width=\textwidth]{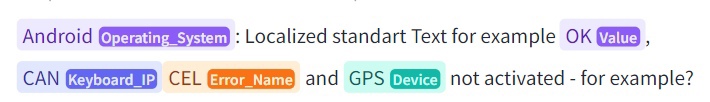}
\caption{Inference of the NER model with wrong broken word CANCEL.} \label{fig2}
\end{figure}

Declared precision of the open-source model is 0.60\footnote{\url{https://huggingface.co/mrm8488/codebert-base-finetuned-stackoverflow-ner}}, hence markup could not be ideal because of model mistakes. Figure~\ref{fig2} shows the Hugging face model inference example. So, annotation models sometimes break a word into several parts and define for each its own class. To smooth out these inaccuracies, we decided to combine parts of words into one entity according to the class of the first defined part. While, entities detected by the model might be ambiguous, testing the key words of sentences mostly results in correct detection. All pairs are vectorized as one-hot encoding, thus each title and question is represented by a k-dimension vector, as there are $k=25$ defined classes.

\subsection{Bayesian Networks}

A Bayesian network is a probabilistic model that encodes a joint probability distribution over a set of variables $V = \{X_1, …, X_n\}$, which, in our case, presents entity classes. We consider only discrete variables. Formally, a Bayesian network $B$ is a pair $\{G, \Theta\}$, where $G$ is a directed acyclic graph called "structure". Each node corresponds to one of the variables from $V$. $\Theta$ is a set of probabilities defined on $G$. It specifies the conditional probability distribution $P(X_i | P A_i)$, where $P A_i$ are the parents of the $X_i$ variable. The lack of edge between the variables encodes the conditional independence. With BN, it is possible to get the joint probability distribution of all variables, given as:
\begin{equation}\label{formula_1}
P(V) = \prod_{i=1}^{n}{P(X_i|P A_i)}
\end{equation}

\subsubsection{Structure Learning}

BNs are a suitable tool for the problem, providing excellent means to structure complex domains and draw inferences. To determine semantic relationships and dependencies, we chose a score-based approach of structure-learning; otherwise, a constraint-based approach needs expert knowledge. In a score-based approach, a scoring function is used to measure how well a given structure fits the data. Formally, the learning problem is to find $B^*$:
\begin{equation}
B^* = \arg\max_{B}{Score(B|D)}
\end{equation}
where $D$ is the given dataset. The score-based approach is essentially a search problem, hence there are two parts: the search algorithm and a score metric.

\subsubsection{Searching Algorithm}


Chickering showed that learning an optimal BN from $D$ is an NP-hard problem \cite{chickering1996learning}. Solving the learning problem precisely becomes impractical, which is why we decided to use the local search algorithm. In our case, the number of variables is equal to 50, because question and title entity classes have 25 each. The search algorithm selected the greedy hill climbing approach \cite{heckerman1998tutorial}. There are also other algorithms that are enabled to learn optimal structure for datasets with dozens of variables\cite{koivisto2004exact, jaakkola2010learning, yuan2012improved}, based on dynamic programming, branch and bound, linear and integer programming (LP), and heuristic search.

\subsubsection{Scoring Metrics}

We used the Bayesian information criterion (BIC) \cite{schwarz1978estimating}, Bayesian Dirichlet equivalent uniform prior(BDeu) and K2 \cite{heckerman1995learning} as metrics. The BIC is based on the Schwarz Information Criterion and consists of a log-likelihood term and a penalty term, defined as $f(X_i, B, D) = log(N)/2$ while the score is defined as follows:
\begin{equation}
Score(B | D) = LL(B | D) - f(X_i, B, D)||V||
\end{equation}
In this way, the influence of model complexity decreases as $||V||$ increases, and we get regularized DAG, as the log-likelihood score usually overfits and tends to favor complete graphs.

BDeu and K2 are scores from the family of Bayesian Dirichlet score functions. Under some assumptions, such as parameter independence, parameter modularity, exchangeable data and Dirichlet prior probabilities it is possible to say that penalty term for BDeu is
\begin{equation}
f(X_i,B,D)=\sum^{q_i}_{j}{\sum^{r_i}_{k}{log{\frac{P(D_{ijk}|D_{ij})}{P(D_{ijk}|D_{ij},a_{ij})}}}},
\end{equation}
where $q_i$ is the number of possible values of $PA_i$, $r_i$ is the number of possible values for $X_i$, $D_{ijk}$ is the number of times $X_i = k$ and $PA_i = j$ in $D$, and $\alpha_{ij}$ is a parameter based on the user-specified $\alpha$. $\alpha$ is a heuristic constant that under the likelihood-equivalent assumption proposes the same distribution, described in general terms by Heckerman, Geiger and Chickering (1995). This is called the equivalent sample size, and low $\alpha$ values typically result in sparse networks. We used $\alpha$ equals to 5, as default value.

After learning structure and finding a local optimum, BNs were pruned by Chi-Square Test Independence \cite{Argyrous1997} to detect more specific semantic relationships.

Additionally, we used the Chow-Liu Algorithm \cite{chow1968approximating}. It finds the maximum-likelihood tree-structured graph (i.e., each node has exactly one parent, except for parentless root node). The score is simply the log-likelihood and there is no penalty term for graph structure complexity as it is regularized by tree structure.

\subsubsection{Predicting \& Evaluating networks}

For BNs using BIC, BDeu and K2 scores, we predicted question' entities using the Maximum Likelihood Estimation (MLE). A natural estimate for the CPDs is to simply use the relative frequencies, with each variable state that has occurred following Formula ~\ref{formula_1}.

For BNs having tree structures we tried different probabilistic inference approaches. Algorithms such as Variable Elimination (VE), Gibbs Sampling (GS), Likelihood Weighting (LW) and Rejection Sampling (RS) are detailed in respective articles \cite{koller2009probabilistic, hrycej1990gibbs}. Each label in question is predicted by a one-vs-rest strategy, by all entities of its title from the pair.

For evaluation we selected common multilabel classification metrics. We preferred macro and weighted averaging because existing classes are imbalanced, and it is important to evaluate each class with its number of instances. The formulas for those metrics are
\begin{equation}
Precision_{M}=\frac{\sum_{i=1}^{k}{\frac{TP_i}{TP_i+FP_i}}}{k};
\end{equation}
\begin{equation}
Precision_{W}=\sum_{i=1}^{k}{Precision_{M_{i}}*W_i},
\end{equation}
where $W_i = \frac{N_i}{N}$, $N_i$  is the number of samples of $i$ class, TP is the number of predictions that correctly reports a positive result, and FP is the number of predictions that incorrectly reports a false positive.

\section{Results}

In this section we analyze classification metrics of BNs based on BIC, BDeu and K2 scores as well as Chow-Liu trees. Each score defines a different structure of DAG, which means different semantic dependencies. We compared DAGs and analyzed the penalty terms of each score and its relationships reflected in graphs, as well as the detected relations.

\subsection{Comparison of Evaluation Metrics}

We used a common train-test split for evaluation. With the dataset described above, we composed the test dataset as random 30\% samples of the whole set. The random seed is defined in a specific way whereby classes from the test set are in the train set as well.

\begin{table}
\caption{Comparison of evaluation metrics.}\label{tab1}
\begin{tabularx}{\textwidth}{|c|*{6}{X|}} \hline
{} & \multicolumn{2}{c|}{Precision} & \multicolumn{2}{c|}{Recall} & \multicolumn{2}{c|}{F1-score} \\ \hline
Model & Macro & Weighted & Macro & Weighted & Macro & Weighted \\ \hline
CatBoost    & 0.41 & 0.58 & 0.19 & \textbf{0.35} & 0.24 & 0.41 \\ \hline
BIC based   & \textbf{0.56} & \textbf{0.66} & 0.20 & 0.33 & 0.28 & 0.42 \\ \hline
BDeu based  & 0.48 & 0.63 & 0.20 & \textbf{0.35} & 0.26 & \textbf{0.43} \\ \hline
K2 based    & 0.51 & \textbf{0.66} & \textbf{0.24} & 0.34 & \textbf{0.29} & \textbf{0.43} \\ \hline
CL trees VE & 0.47 & 0.63 & 0.21 & 0.33 & 0.25 & 0.41 \\ \hline
CL trees LW & 0.48 & 0.63 & 0.17 & 0.29 & 0.22 & 0.37 \\ \hline
CL trees GS & 0.41 & 0.57 & 0.13 & 0.25 & 0.18 & 0.33 \\ \hline
CL trees RS & 0.23 & 0.44 & 0.07 & 0.15 & 0.10 & 0.22 \\ \hline
\end{tabularx}
\end{table}

Table \ref{tab1} shows the main evaluation results according to the selected classification metrics. We prefer to accentuate precision, because precision of individual classes is most important for information extraction and context prediction, and wrong class predictions caused context misunderstanding.

Our approach shows better precision metrics than the baseline - CatBoost model \cite{NEURIPS2018_14491b75}, 0.56 vs 0.41 macro precision and 0.66 vs 0.58 weighted precision, comparing the BIC score-based network and baseline.

We observe the highest precision in the BIC score-based model, while the K2-based model shows the better recall metric and comparable precision, hence the best network from the F1-score perspective is the K2-based one. As expected, the BIC regularizes the log-likelihood stronger than the BDeu and K2-specific penalty terms. As a result, BDeu and K2-based DAGs detect more relationships that allow to classify more instances of each class correctly, hence the growth of recall.

We see that the Chow-Liu tree-based networks are comparable to other models if Variable Elimination is used as a sampling algorithm. This causes the limitation that each node has exactly one parent, except parents root nodes, and it is non-redundant for the DAG to fit the data. Other sampling algorithms approximating solutions to the inference problem show worse results.

\subsection{Visual DAG representation}

We visualized DAGs from each Bayesian network to see relationships that a BN allows to detect. A BN has causal structure, and we use this property to analyze connections of different semantic entities, describing the context. Figure~\ref{fig:dags} shows structures learned by described methods. The graphs provide information about the relations between significant parts of the context.

As expected, graphs of K2 (\ref{fig:k2}) and BDeu (\ref{fig:bdue}) -based networks detect more relationships and are more complete, in contrast to the BIC (\ref{fig:bic}) -based graph. At each DAG there are semantic links between the same title and question entity classes. The structure of the Chow-Liu trees (\ref{fig:cl}) shows this very well.

Additionally, analysis shows different clusters of semantic entities. This way, DATA STRUCTURE and ALGORITHM separated in each of the four graphs. Furthermore, there is a link between FILE NAME and FILE TYPE, CODE BLOCK and OUTPUT BLOCK. This indicates the logic and validity of BN DAGs structures.

It is noteworthy that the tree structured DAG defines causation from Question ALGORITHM to Title USER NAME and from Title ALGORITHM to Question CLASS without establishing a causal relationship between entities of the same names. More likely, these are outliers, as the NER model is not ideal.

\begin{figure}
\centering
\begin{subfigure}{0.85\textwidth}
    \includegraphics[width=\textwidth]{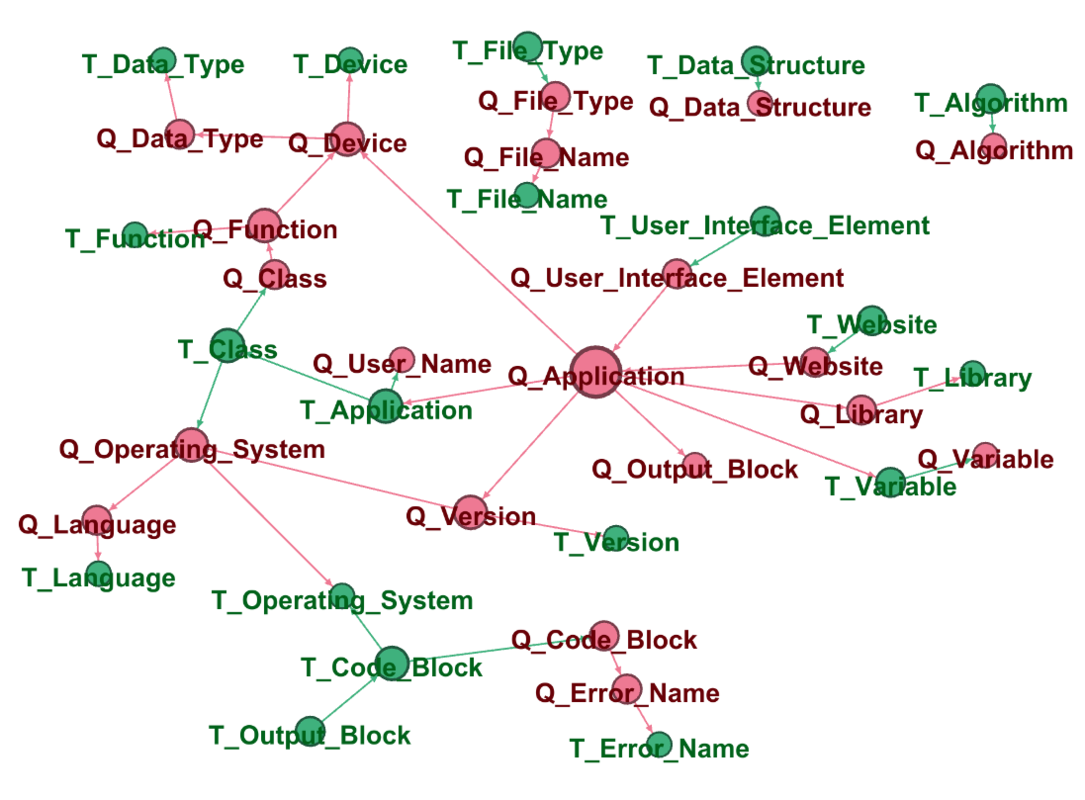} 
    \caption{BIC based graph}
    \label{fig:bic}
\end{subfigure}
\hfill
\begin{subfigure}{0.85\textwidth}
    \includegraphics[width=\textwidth]{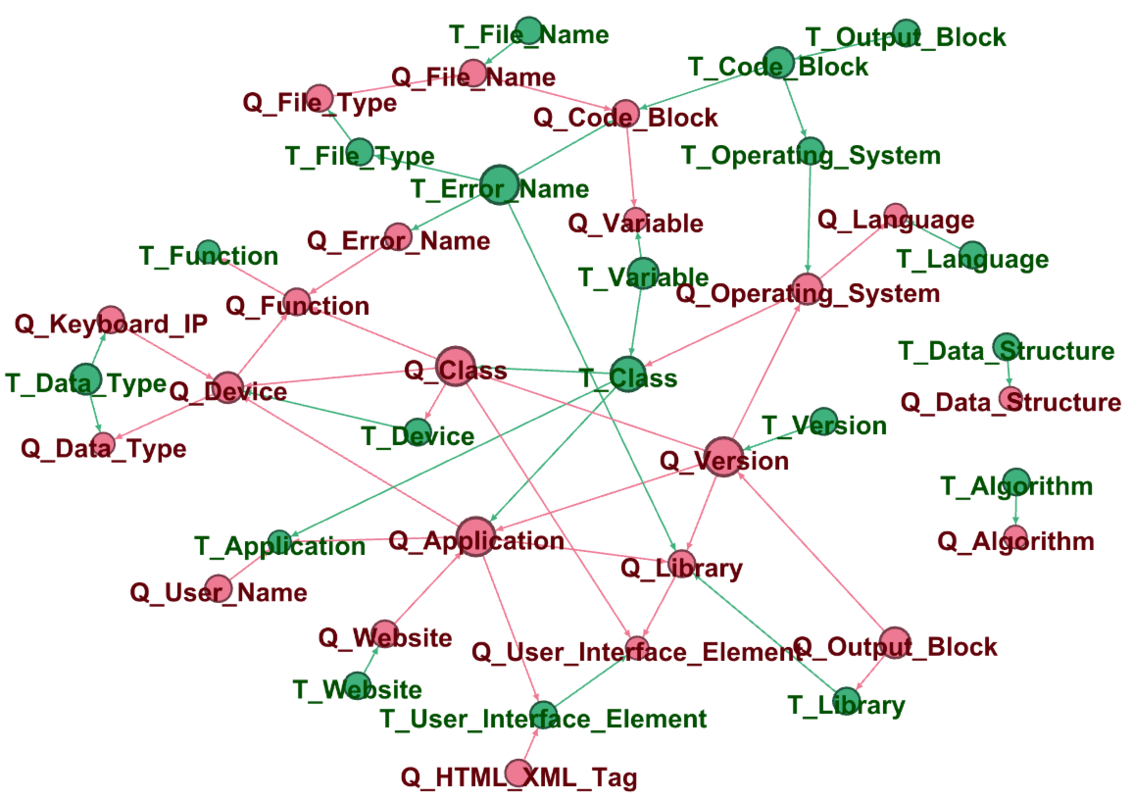}
    \caption{K2-metric based graph}
    \label{fig:k2}
\end{subfigure}
\end{figure}

\begin{figure}\ContinuedFloat
\centering
\begin{subfigure}{0.85\textwidth}
    \includegraphics[width=\textwidth]{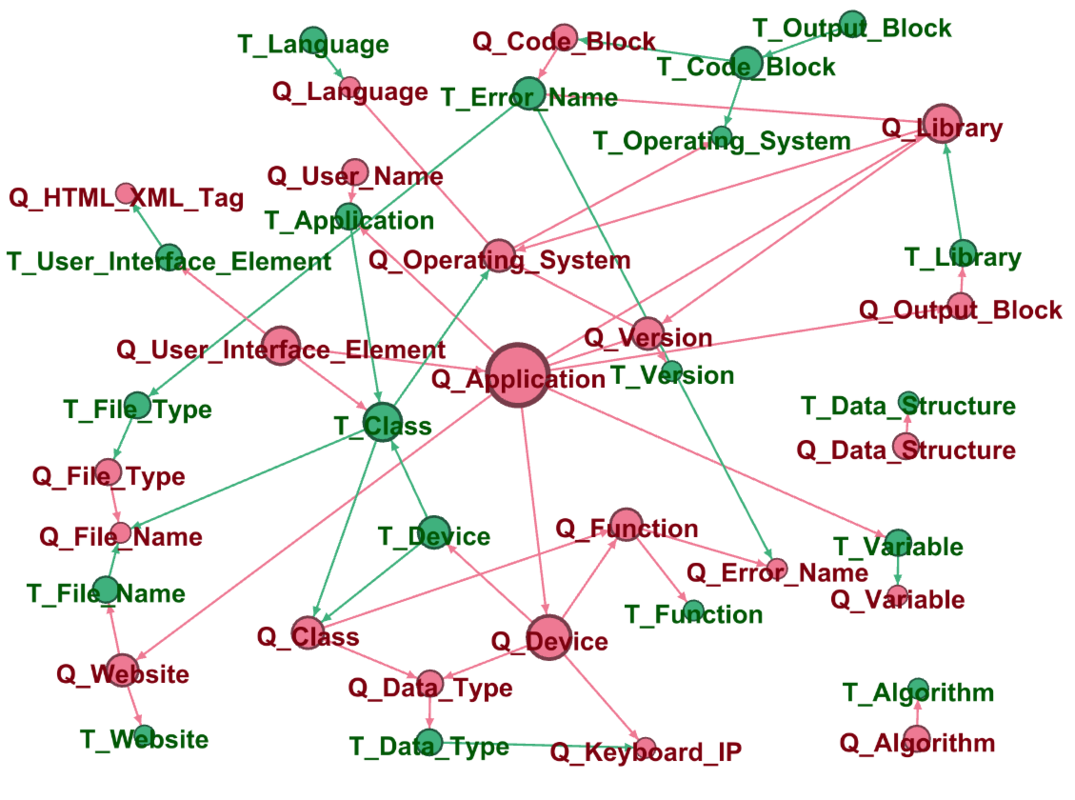}
    \caption{BDeu based graph}
    \label{fig:bdue}
\end{subfigure}
\hfill
\begin{subfigure}{0.85\textwidth}
    \includegraphics[width=\textwidth]{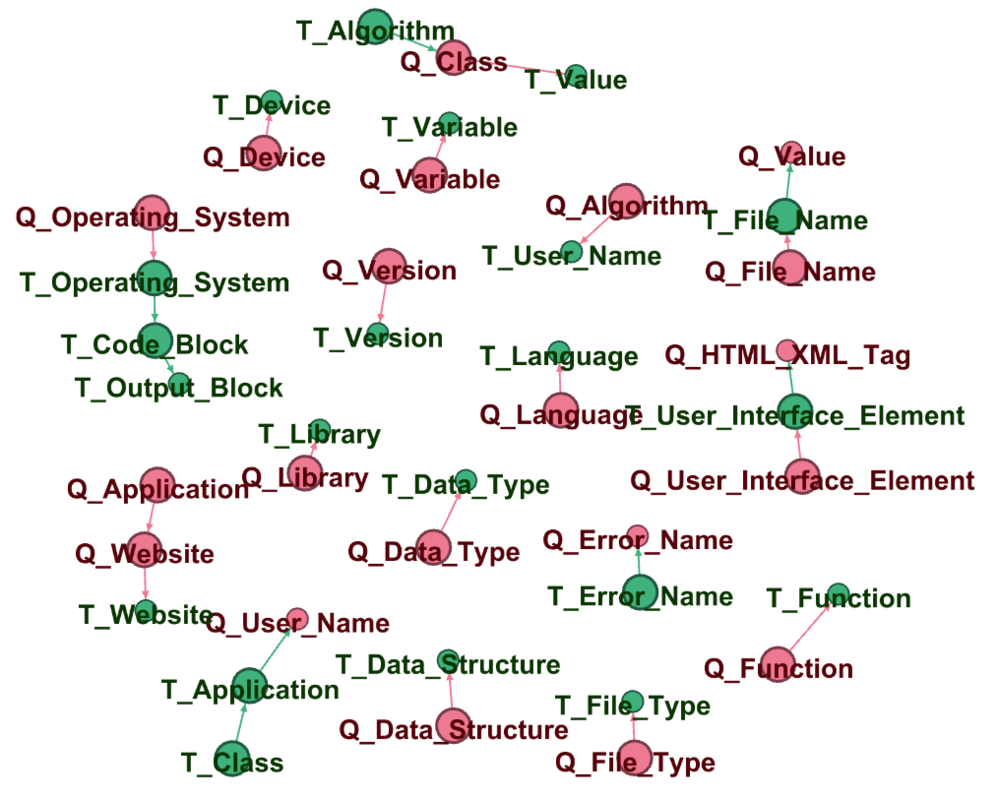}
    \caption{Chow-Liu trees based graph}
    \label{fig:cl}
\end{subfigure}
\hfill
\caption{DAG structures of learned BNs}
\label{fig:dags}
\end{figure}

\subsection{Predictions analysis}

Finally, we compared semantic entities detected by the NER model and predicted by BN based on the K2 metric. Below in Table~\ref{tab2}, there are several examples of predictions. Matched to the DAG described above, we observe that overall, predicted entities coincide to target one in the first example. In some cases, BN could not detect semantic instances, as in the second and third rows of Table~\ref{tab2}. On Graph (\ref{fig:k2}) VERSION and USER NAME have consequence relations from APPLICATION at both question and title. Similarly, OPERATING SYSTEMS connected to APPLICATION and LANGUAGE in the graph. It is likely that the value of conditional probability was not enough to consider whether these entities would be in the question context.

\begin{table}
\caption{Comparison of existing and predicted entities.}\label{tab2}
\begin{tabular}{|p{2cm}|p{4.8cm}|p{2.5cm}|p{2.5cm}|} \hline
{Title} & {Question} & {Questions entities} & {Predicted entities} \\ \hline
{How to send email with attachment using GmailSender in android} & {I want to know about how to send email with attachment using GmailSender in android.} & {APPLICATION, OPERATING SYSTEM} & {APPLICATION, OPERATING SYSTEM} \\ \hline
{Intel XDK build for previous versions of Android} & {I have just started developing apps in Intel XDK and was just wondering how to build an app for a specific version of Android OS. The emulator I select "Samsung Galaxy S" is using the version 4.2 of android. My application works fine for Galaxy s3 but not on galaxy Ace 3.2 . I could not find a way to add more devices to the emulator list. How can I achieve this. Regards, Shankar.} & {APPLICATION, OPERATING SYSTEM, VERSION, USER NAME} & {APPLICATION, OPERATING SYSTEM} \\ \hline
{Automatic update database of android application} & {I'm making an quiz application in android.But If there are changes in database then how can user get updated with this changes.I read about GCM and php.But can anyone tell me how to do that?Any helpful tutorial? Thanks.} & {OPERATING SYSTEM, APPLICATION, LANGUAGE} & {OPERATING SYSTEM} \\ \hline
\end{tabular}
\end{table}

\section{Discussion}

As mentioned above, Bayesian networks are one of the methods to predict and analyze the context. This method may be especially useful at the CQA domain for information extraction and semantic causation in the analysis of the important parts of a question and how clear it is for answering \cite{9892454}. Results show that BN are able to capture the main trend of using meaningful entities, in particular in the programming domain. The recovery task might be an efficient way to determine heuristics as an improvement of the BN approach for context prediction and meaning, as shown by Mehmet et al. Global Uniform parameter priors \cite{10.5555/2073876.2073906}, because we have no knowledge about prior distribution. Conversely, an additional penalty term could fare better in structure learning and detect more relevant relations. Finally, using the mentioned optimal search algorithm should show better defined metrics.

Another approach is work with data. On the one hand, expanding data and not specifying the android tag could allow the BN to determine more general dependencies because of Bayesian inference. It is also possible to change data representation, and focus on specific meaningful words or verbal constructions as opposed general classes of entities. It could lead to the growth of variables $V$ power depending on penalty term structure capacity, due to context uncertainty.

Furthermore, there are techniques of query expansion based on relevant documents feedback, especially in information retrieval systems \cite{kandasamy}. Neural systems lack interoperability, whereas Bayesian networks have a clarify causal inference and could potentially be a good tool for query expansion and reformulations by providing context representation from the given query-reformulation pairs \cite{adolphs}.

\section{Conclusion and Future Directions}

In this paper we proposed a new application for Bayesian networks in CQA. Bayesian networks could be used as a tool for context prediction and context information extraction. Applying BN to CQA and programming domain in this way, we recognized causal semantic relationships on the set of SO questions and related titles. More precisely, we received the DAGs based on different approaches, making it possible to analyze interrelations. Moreover, we defined that BNs identify entities acceptably, mostly correctly but with issues detecting semantic classes that are separated in the DAG structure.

In future work we plan to build an end-to-end artificial neural network based on the existing NER model. In addition, it seems interesting to compare the NER model and the Bayesian network approach on a small dataset, as done in this article. With this in mind, we would use LSTM with attention to predict semantic entities. Additionally, we are planning to compare the BN and LDA (Latent Dirichlet allocation) approaches for the problems of thematic modeling and information extraction CQA.

\bibliographystyle{splncs03_unsrt}
\bibliography{biblio}

\end{document}